%% file: main.tex
\documentclass{article} %
\usepackage{iclr2023_conference,times}

\input{math_commands.tex}

\usepackage{hyperref}
\usepackage{url}

\usepackage{booktabs}       %
\usepackage{amsfonts}       %
\usepackage{amsmath}
\usepackage{xcolor}         %

\usepackage{multirow}
\usepackage{multicol}
\usepackage{hhline}
\usepackage{xspace}
\usepackage{adjustbox}
\usepackage{import}
\usepackage{subcaption}
\usepackage{color, colortbl}
\usepackage{makecell}
\usepackage{wrapfig}
\usepackage{newtxmath}
\usepackage{authblk}

\subimport{./}{macro}

\title{\ours: Multi-attribute Selective Suppression}

\author[1]{\bf{Chun-Fu (Richard) Chen}}
\author[1]{\bf{Shaohan Hu}}
\author[ ]{\bf{Zhonghao Shi}$^{1, 2}$\thanks{Work done during  internship at JPMorgan Chase Bank, N.A.}}
\author[ ]{\bf{Prateek Gulati}$^{1, 3*}$}
\author[1]{\\ \bf{Bill Moriarty}}
\author[1]{\bf{Marco Pistoia}}
\author[4]{\bf{Vincenzo Piuri}}
\author[4]{\bf{Pierangela Samarati}}
\affil[1]{JPMorgan Chase Bank, N.A., USA}
\affil[2]{University of Southern California, USA}
\affil[3]{Northeastern University, USA}
\affil[4]{Universit\`{a} degli Studi di Milano, Italy}
\affil[ ]{ }
\affil[ ]{\small\{richard.cf.chen, shaohan.hu, william.r.moriarty, marco.pistoia\}@jpmchase.com,}
\affil[ ]{\small zhonghas@usc.edu, gulati.p@northeastern.edu,}
\affil[ ]{\small\{vincenzo.piuri, pierangela.samarati\}@unimi.it}

\iclrfinalcopy %
\begin{document}

\maketitle

\begin{abstract}

The recent rapid advances in machine learning technologies largely depend on the vast richness of data available today, in terms of both the quantity and the rich content contained within. For example, biometric data such as images and voices could reveal people's attributes like age, gender, sentiment, and origin, whereas location/motion data could be used to infer people's activity levels, transportation modes, and life habits. Along with the new services and applications enabled by such technological advances, various governmental policies are put in place to regulate such data usage and protect people's privacy and rights. As a result, data owners often opt for simple data obfuscation (e.g., blur people's faces in images) or withholding data altogether, which leads to severe data quality degradation and greatly limits the data's potential utility.

Aiming for a sophisticated mechanism which gives data owners fine-grained control while retaining the maximal degree of data utility, we propose \textit{Multi-attribute Selective Suppression}, or \textit{\ours}, a general framework for performing precisely targeted data surgery to simultaneously \textit{suppress} any selected set of attributes while \textit{preserving} the rest for downstream machine learning tasks. \ours learns a data modifier through adversarial games between two sets of networks, where one is aimed at suppressing selected attributes, and the other ensures the retention of the rest of the attributes via general contrastive loss as well as explicit classification metrics. We carried out an extensive evaluation of our proposed method using multiple datasets from different domains including facial images, voice audio, and video clips, and obtained promising results in \ours' generalizability and capability of suppressing targeted attributes without negatively affecting the data's usability in other downstream ML tasks.

\end{abstract}

\subimport{./}{Tex/01_introduction}

\subimport{./}{Tex/02_related_works}

\subimport{./}{Tex/03_method}

\subimport{./}{Tex/04_experiments}

\subimport{./}{Tex/05_conclusion}

\paragraph{Code of Ethics and Ethics statement.} 
Our work is to selectively suppress the attributes in multi-attribute data while preserving their potential utility
such that the sensitive information in the data could be minimized and hence the data could be used to benefit the community.
We believe that there are no ethical concerns related to this work.

\paragraph{Reproducibility Statement.} We provided the training and evaluation details in the main paper and appendix. 
Our source codes and models will be publicly available to help better understand the settings of training and evaluation.

\section*{Disclaimer}
This paper was prepared for information purposes by the
teams of researchers from the various institutions identified
above, including the Global Technology Applied Research
group of JPMorgan Chase Bank, N.A.. This paper is not a product of the Research Department of JPMorgan Chase Bank, N.A. or its affiliates. Neither JPMorgan Chase Bank, N.A. nor any of its affiliates make any explicit or implied representation or warranty and none of them accept any liability in connection with this paper, including, but limited to, the completeness, accuracy, reliability of information contained herein and the potential legal, compliance, tax or accounting effects thereof. This document is not intended as investment research or investment advice, or a recommendation, offer or solicitation for the purchase or sale of any security, financial instrument, financial product or service, or to be used in any way for evaluating the merits of participating in any transaction.

\clearpage
\bibliography{iclr2023_conference}
\bibliographystyle{iclr2023_conference}

\appendix
\subimport{./}{Tex/09_appendix}

\end{document}

%% file: math_commands.tex
\usepackage{amsmath,amsfonts,bm}

\def\eqref#1{equation~\ref{#1}}

\def\1{\bm{1}}

\DeclareMathAlphabet{\mathsfit}{\encodingdefault}{\sfdefault}{m}{sl}
\SetMathAlphabet{\mathsfit}{bold}{\encodingdefault}{\sfdefault}{bx}{n}

\newcommand{\KL}{D_{\mathrm{KL}}}

%% file: macro.tex
\graphicspath{{ICLR2023/Main/}}

\def\ours{MaSS\xspace}
\def\adienceid{DataID\xspace}
\def\audiomnistid{SpeakerID\xspace}

\newcommand{\tablevspace}{\vspace{-8pt}}
\newcommand{\figurevspace}{\vspace{-6pt}}

\definecolor{Gray}{gray}{0.9}

%% file: Tex/01_introduction.tex
\section{Introduction}
\label{sec:intro}

The recent rapid advances in machine learning (ML)
can be largely attributed to powerful computing infrastructures
as well as the availability of large-scale datasets, 
such as ImageNet1K~\citep{deng2009imagenet} for computer vision, 
WMT~\citep{wmt19translate} for neural machine translation, and LibriLight~\citep{librilight} for speech recognition.
Studies have shown that 
ML models trained on large-scale datasets
can usually prove effective in
many additional downstream tasks~\citep{fewshotlearner}.
On the other hand, 
ethical concerns have been raised surrounding proper data usage
in issues like data privacy~\citep{data_privacy}, data minimization~\citep{data_minimization}, etc.
Therefore, if there are more data available and can be used without worrying whether or not the data is handled properly, 
the ML models can be further improved by more data and help the ML community to advance on many domains.

Attempting to balance between model performance
and proper data usages,
a common approach usually taken
is to simply modify the data to remove
its ``sensitive'' attributes,
and experimentally demonstrate 
that the targeted sensitive attributes
are indeed removed.
What's crucially important but usually omitted 
here,
however,
is the preservation of the ``total utility'' of the data,
because the suppression operation oftentimes also
negatively impact, or even completely destroy,
the other ``non-sensitive'' attributes,
hence greatly damaging the dataset's potential future utility.
For example, DeepPrivacy~\citep{deepprivacy}
\textit{is} able to demonstrate its privacy protection capability,
but the modified data it produces can no longer
be utilized for additional downstream tasks like
sentiment analysis, age detection, or gender classification.
Since data is one of the main driving forces
for the rapid advancement of machine learning research,
we argue that the ideal scenario would be
to have the flexibility of selecting an arbitrary set
of attributes and only suppressing \textit{them}
while leaving all the other attributes completely intact.
In this way, the community could unleash the potential utility of the modified data to develop more advanced algorithms.

\begin{figure}[t]
    \centering
    \includegraphics[width=0.95\linewidth]{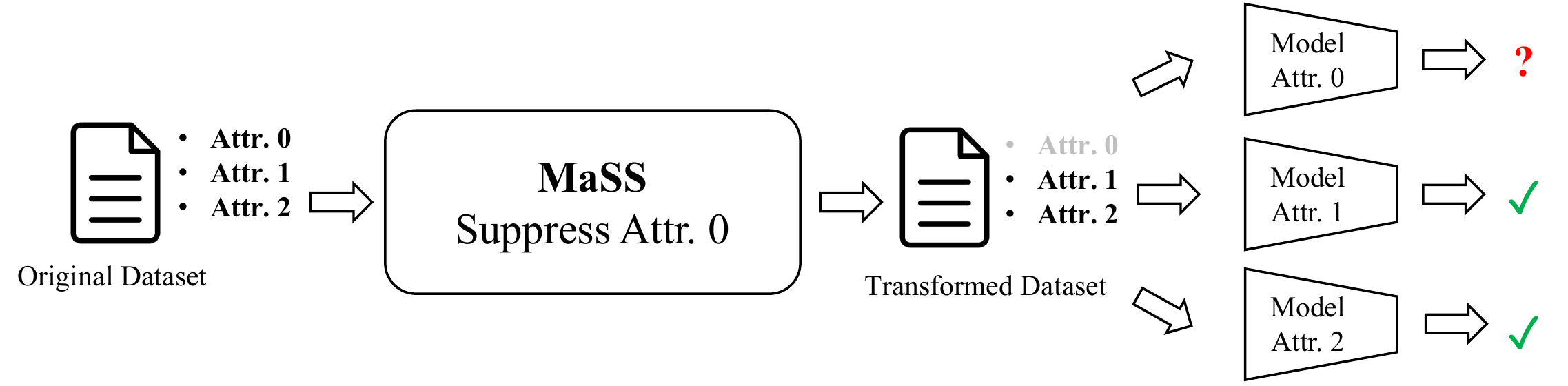}
    \figurevspace
    \caption{
    \ours is able to precisely target any selected attributes in a multi-attribute dataset for suppression
    while leaving the rest of the
    attributes intact
    for any potential downstream ML-based analytic tasks.
    For example as illustrated by the diagram,
    when operating on the original multi-attribute 
    dataset and configured to 
    suppress Attr. 0,
    \ours is able to transform the dataset
    such that the model for detecting Attr. 0 
    is unable to reliably detect Attr. 0 from the transformed data, while the models for Attr. 1 and 2 are not affected.
    }
    \label{fig:intro}
    \vspace{-6pt}
\end{figure}

Towards this exact goal, we present
Multi-attribute Selective Suppression (or \ours) in this paper,
to enable such capability of precise attributes suppression 
for multi-attribute datasets.
The high-level objective of \ours
is also illustrated in Figure~\ref{fig:intro},
where \ours is configured to suppress Attr. 0 without knowing in advance that Attr. 1 and 2 
will be used for downstream tasks.
After the data transformation performed by \ours, 
Attr. 0 becomes suppressed,
nondetectable by its corresponding machine learning model,
but at the same time,
Attr. 1 and 2 are left intact,
and still can be extracted by their corresponding ML models.
As a concrete example, 
suppose we are working with a facial-image dataset
which contains attributes like age, gender, and sentiment,
where, let us assume, age and gender are considered sensitive.
Then, \ours would transform
this facial-image dataset 
such that age and gender information could no longer be 
not be inferred by the corresponding ML models,
whereas sentiment information could still be extracted from the transformed data.

The contributions of our work are threefolds,
\begin{enumerate}
    \item We propose the novel \ours framework to 
    enable the powerful flexibility of precise suppression 
    of arbitrary, selective data attributes.
    \item We employ multiple learning mechanisms in \ours 
    to enable its attribute-specific as well as
    generic feature preservation capabilities,
    which help it achieve satisfactory data utility protection
    both with and without the prior knowledge about downstream tasks.
    \item We thoroughly validate \ours using a wide range of multi-attribute datasets, including image, audio, and video. 
    All our results demonstrate \ours' strong performance in its intended selective attribute suppression and preservation.
\end{enumerate}

%% file: Tex/02_related_works.tex
\section{Related Works}
\label{sec:related_works}

\paragraph{Data Privacy.} 
A large body of work has studied methods applying generative adversarial networks (GANs) to generate and modify facial features in images, so these identity-related sensitive features can be de-identified.
DeepPrivacy~\citep{deepprivacy} proposed to use a conditional generative adversarial network to generate realistic anonymized faces, while considering the existing background and a sparse pose annotation. 
To further ensure the face anonymization using GAN-based methods,
CIAGAN~\citep{CIAGAN_Maximov_2020_CVPR} proposed an identity 
control discriminator to control which fake identity is used 
in the anonymizaiton process by introducing an identity control vector. 
Instead of generating the entire faces for anonymization, 
\citeauthor{li2021identity} proposed to apply conditional 
GAN to only identify and modify the five identity-sensitive 
attributes. To also enable face anonymization with the selected 
semantic attributes manipulation, PI-Net~\citep{chen2021perceptual} 
proposed to generate realistic looking faces with the selected 
attributes preserved.
The above works usually focuses on suppression only while the future data utilities are not considered. 
Our approach not only suppresses the attributes but also preserves the data utility concurrently.
On the other hand, SPAct tried to suppress the multiple attributes in a video through contrastive learning while preserving the utility for action recognition; 
however, their approach lacks the flexibility to handle individual attributes but can only process all attributes at once and limits to the action recognition dataset; while our method is fully configurable and validated in different data domains. 
\citeauthor{moriarty2022utility} proposed the method to suppress the biometric information while preserving its utility; 
however, their approach requires the information of downstream task while our method does not.

\paragraph{Dataset Distillation/Condensation.} 
Dataset distillation and condensation
aim to create a smaller version of a large dataset which can be used to train 
a model whose performance can be as good as training on the original 
large dataset.
Therefore, training can be relatively quicker, 
e.g., Neural Architecture Search (NAS) methods require a lot of 
iterations of a whole dataset to find out the best model. Otherwise, 
NAS usually needs to use a proxy dataset/model for the approximation 
results from large dataset/model~\citep{zhao2021datasetcondensation, Cazenavette_2022_CVPR_Dataset_Distillation, Wang_2022_CVPR_cafe}.
On the other hand, a recent work shows the condensed dataset also conceals 
some attributes from the original data~\citep{pmlr-v162-dong22c-condensation-privacy}
but still remains effective for the original task.
In contrast to these approaches, our proposed method tries to keep the truthfulness of data as much as possible, so we do not reduce the amount of data.
Moreover, our method is designed to preserve the generic features rather than the specific task, which could be covered by the generic feature we preserved.

\paragraph{Self-supervised/Contrastive Learning.}
Self-supervised learning has been explored by the community to allow a machine learning model to learn a good data representation by designing pretext tasks instead of human annotations~\citep{rotnet_selfsupervised}, or contrastive learning which tries to maximize the agreement between positive pairs~\citep{pmlr-v119-chen20j-simclr,he2020momentum,grill2020bootstrap}, or clustering-based methods to generate pseudo labels for data~\citep{caron2020unsupervised,Caron_2018_ECCV_deepclustering}, or mask autoencoder to predict the masked patches by the remaining patches~\citep{he2022masked}.
Then, they show that the feature representations are usually good for many different downstream tasks.
Our goal is to keep as much information when suppressing the selected attributes without having prior knowledge; thus, self-supervised and contrastive learning methods facilitate our requirement to extract generic features without label information
from downstream tasks.

%% file: Tex/03_method.tex
\section{Proposed Method}
\label{sec:proposed_methods}

In this section
we give a formal description of our target problem,
and discuss our proposed \ours framework
and all its core components in detail.

\sloppy

\subsection{Problem Definition}
Consider a multi-attribute dataset $\mathbf{X}$ with size $N$,
and the set $A$ of different attributes,
where each data point $\mathbf{x}$'s value for attribute $a \in A$ is $a_{\mathbf{x}}$.
Each of the different attributes can be learned 
by inferencing on $\mathbf{X}$,
hence our objective is to transform $\mathbf{X}$ 
to suppress any selective subset of attributes $S \subseteq A$
such that no attributes in $S$ can be reliably inferred from the transformed dataset $\mathbf{X}'$.
At the same time, we need to make sure 
the rest of the attributes $R=A \setminus S$ are preserved
and can still be inferred from $\mathbf{X}'$.

In practice, when a data owner would like to transform their data,
we assume the subset $S$ of attributes targeted for suppression is always predetermined.
However, it is not always known in advance
what the entire set $A$ of attributes are,
and consequently which set $R$ of attributes need to be preserved.
Therefore, for generalizability, we consider the case of an unknown $R$ at the time of data transformation.
Of course, if $R$ happens to be given, our solution needs to be able to take advantage of this extra information as well.

\subsection{Proposed Framework}

\begin{figure}[bt]
    \centering
    \includegraphics[width=\linewidth]{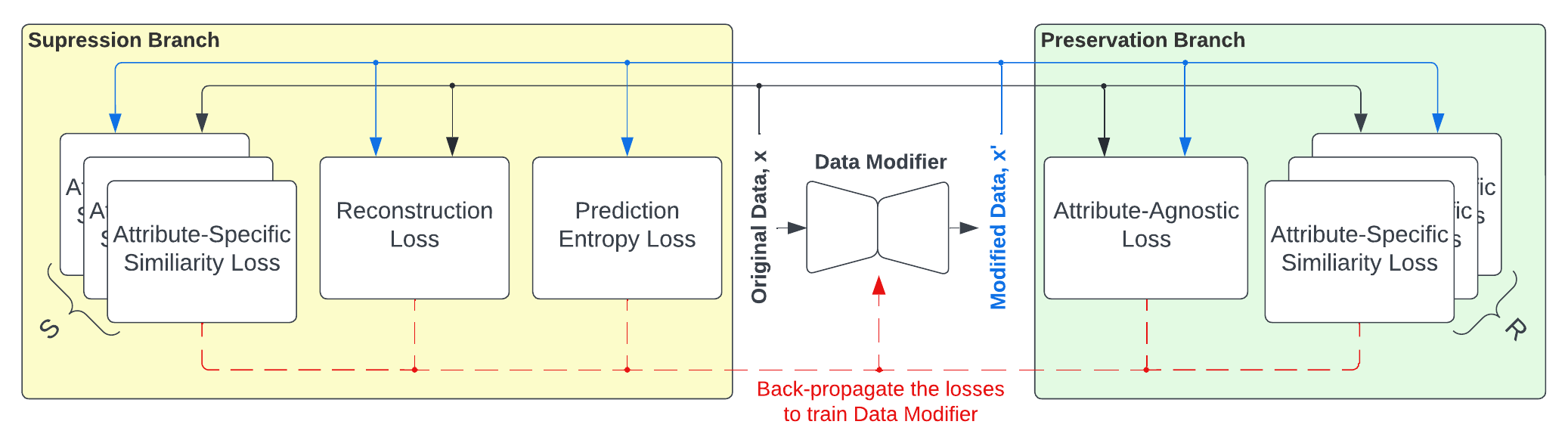}
    \figurevspace
    \caption{
    \textbf{Overview of \ours .}
    \ours contains three components: the data modifier, the suppression branch, and the preservation branch. 
    The data modifier $G$ is trained by optimizing the losses in both the suppression and the preservation branches.
    }
    \label{fig:framework}
\end{figure}

We propose a Generative Adversarial Network (GAN)-based solution in tackling
the data transformation problem.
Our framework consists of three major components: 
a \textit{Data Modifier},
a \textit{Suppression Branch}, and
a \textit{Preservation Branch},
as depicted in Fig.~\ref{fig:framework}.
The data modifier $G$ is the generator while both branches are the discriminators 
in the GAN framework.
The data modifier tries to generate new data such that
the similarity between original data and modified data are maximized and minimized via the suppression branch and the preservation branch, respectively.
In a nutshell, the data modifier learns a transformation
that is to be applied to the original data vectors,
where the learned transformation is jointly regularized by both
the suppression and preservation branches
to ensure all targeted attributes in $S$ are indeed suppressed in the transformed data,
while all other attributes, explicitly specified or not,
are left intact as much as possible.
We next discuss all three components in more detail.

\subsubsection{Data Modifier}

We design the data modifier to be conditional on its original input in learning instance-specific transformations
via a multi-layer perceptron (MLP) with a residual shortcut.
In other words, based on its input $\mathbf{x}$, the data modifier $G$ learns an additive modification to be applied to $\mathbf{x}$. 
Therefore, the transformed data stays in the same embedding space as the original, which leads to faster optimization convergence.
Specifically, for a normalized original data vector $\mathbf{x}$, its normalized transformed version is computed as
\begin{equation}
\label{eq:g_func}
    \mathbf{x}' = G(\mathbf{x}) := n(\mathbf{x} + n({\text{MLP}(\mathbf{x})})),
\end{equation}
where $n(\cdot)$ is the normalization function $n(\mathbf{v}) = 
\begin{cases}
\frac{\mathbf{v}}{||\mathbf{v}||}, & ||\mathbf{v}|| \neq 0\\
0, & \text{o.w.}
\end{cases}
$.

During the optimization, the parameters in the data modifier are trained by back-propagating the losses designed in both branches.

\subsubsection{Suppression Branch}

The suppression branch is designed to make the targeted attributes in $S$ as unrecognizable as possible.
It utilizes the corresponding set of inference models 
pretrained on the original dataset $\mathbf{X}$.
Each pretrained model 
corresponds to a specific attribute $s \in S$
and is composed of 
a feature extractor $F^{s}$, which converts raw input data to a feature vector
$\mathbf{z}_{\mathbf{x}}^{s} = F^{s}(\mathbf{x})$,
and a projector $P^{s}$, which maps the features to the attribute's label
$\mathbf{p}_{\mathbf{x}}^{s} = P^{s}\left(\mathbf{z}_{\mathbf{x}}^{s}\right)$,
where $\mathbf{z}_{\mathbf{x}}^{s}$ is the feature representation and $\mathbf{p}_{\mathbf{x}}^{s}$ is the prediction logit.

During the training of the data modifier, 
the suppression branch guides the data modifier 
to degrade these pretrained models' recognition accuracies 
on the transformed data $\mathbf{X}'$.
It does so by either measuring the similarity of features from the pretrained models 
or by comparing the prediction results against the ground truth labels, 
and then adding a corresponding penalty.
Specifically, for each targeted attribute $s \in S$, 
the feature-similarity loss function is defined as
\begin{equation}
\label{eq:suppression_loss}
    L^{s}_\text{sim} = w^{s} \cdot \frac{1}{N}\sum_{\mathbf{x} \in \mathbf{X}}{
        \text{sim}^{s}\left(
            \mathbf{x}',
            \mathbf{x}
        \right)
    },
\end{equation}
where $w^{s}$ is the weight for attribute $s$ 
and $\text{sim}^{s}(\cdot, \cdot)$ defines a similarity measure.
For example, we can use the cosine similarity between feature vectors
$\text{sim}^{s}_{\cos}\left(\mathbf{x}', \mathbf{x}\right)
= \cos\left(\mathbf{z}_{\mathbf{x}'}^{s}, 
              \mathbf{z}_{\mathbf{x}}^{s}\right)$,
or the negative KL-divergence between the original and the transformed logits
$\text{sim}^{s}_{\tiny\mbox{KL}}\left(\mathbf{x}', \mathbf{x}\right)
= -\KL \left(\mathbf{p}_{\mathbf{x}'}^{s} \Big\Vert\ 
               \mathbf{p}_{\mathbf{x}}^{s}\right)$,
or the cross-entropy loss between the transformed logits 
and the attribute's ground truth value
$\text{sim}^{s}_{\tiny\mbox{CE}}\left(\mathbf{x}',\mathbf{x} \right)
= -\KL \left(\mathbf{p}_{\mathbf{x}'}^{s} \Big\Vert\ 
               \mathbf{1}_{s_{\mathbf{x}}}\right)$,
where $s_{\mathbf{x}}$ is $\mathbf{x}$'s ground truth value for attribute $s$ 
and $\mathbf{1}_{s_{\mathbf{x}}}$ is the corresponding one-hot vector.

The above similarity-based metrics only help guide the data modifier
to lead the pretrained models towards incorrect predictions about $S$ on the transformed data.
However, in terms of attribute suppression, 
an even stronger condition would be to reduce such predictions to random guesses.
Therefore, we introduce an additional loss term 
to maximize the entropy $H$ of the predictions about $S$ on the transformed data
such that the predictions look like random guesses.
Combined with Eq.~\ref{eq:suppression_loss}, the total loss on the suppression branch is
\begin{equation}
\label{eq:suppression_loss_all}
    L^{s} =   w^{s} \cdot \frac{1}{N}\sum_{\mathbf{x} \in \mathbf{X}}{
                    \text{sim}^{s}\left(
                        \mathbf{x}',
                        \mathbf{x}
                    \right)
                } -
                h^{s} \cdot \frac{1}{N}\sum_{\mathbf{x} \in \mathbf{X}}{
                    H\left(
                        \mathbf{p}_{\mathbf{x}'}^{s}
                    \right)
                },
\end{equation}
where $h^{s}$ is the weight for the entropy loss.

Moreover, to regularize the learned data transformation to be as small as possible, we also add an $L$2 reconstruction loss term in the suppression branch
\begin{equation}
\label{eq:rec_loss}
    L_{\text{rec}} = w_{\text{rec}} \cdot \frac{1}{N}\sum_{\mathbf{x} \in \mathbf{X}}{||\mathbf{x}' - \mathbf{x} ||_2}.
\end{equation}

\subsubsection{Preservation Branch}

While the suppression branch guides the data modifier to decrease the confidence of machine learning models on certain attributes $S$ from the data,
the preservation branch is responsible for guarding the data against said suppression and erasure
such that maximum utility can be preserved through the transformation,
in the sense that all the attributes $R = A \setminus S$ not targeted by the suppression branch
should remain recognizable from the transformed data $\mathbf{X}'$,
just as they were from the original data $\mathbf{X}$.
In the proposed method, we design two types of losses: one is \textit{attribute-agnostic} in the preservation branch,
which is applicable to any attribute as the loss is defined in an agnostic way; 
another one is \textit{attribute-specific}, which if the downstream tasks have been defined and we know the set of $R$, then we embedded them into the loss.

\paragraph{Attribute-Agnostic.}
As previously discussed,
it is oftentimes difficult to identify in advance the set $A$ of \textit{all} attributes associated with a dataset,
which means the set $R$ might not necessarily be defined even when $S$ is specified.
The absence of $R$ indicates that we have no knowledge about what attributes downstream tasks might want to detect from the transformed dataset.

To tackle this \textit{attribute-agnostic} scenario, 
we employ self-supervised techniques,
where a generic feature representation of the data is learned without any specific attribute or task information.
For our preservation branch in particular,
we adopt the SimCLR~\citep{pmlr-v119-chen20j-simclr}
contrastive-learning-based approach
to pretrain an extractor $F^{r_*}$ and a projector $P^{r_*}$
that maximize the similarity between embedding pairs originated from the same data point
(i.e., positive pairs)
and minimize that of different data points (i.e., negative pairs).
Instead of considering two different transformations like SimCLR does,
we treat an original data point $\mathbf{x}$ and its transformed version $\mathbf{x}'$
as a positive pair,
and all the rest
as negative.
We enforce this relationship by applying normalized temperature-scaled cross entropy (NT-Xent) loss function introduced in SimCLR; therefore, semantically, the transformed data could preserve more generic features.
For the original data point $\mathbf{x}$-based positive pair $(\mathbf{x}, \mathbf{x}')$, the NT-Xent loss $l_\mathbf{x}^{r_*}$ is
\begin{equation}
    \label{eq:simclr0}
    l_\mathbf{x}^{r_*} = -\log\frac{
        e^{
            \cos\left(
                \mathbf{p}_\mathbf{x}^{r_*}, 
                \mathbf{p}_{\mathbf{x}'}^{r_*}
            \right) / \tau
        }
    }
    {
        \sum_{\mathbf{y} \in \mathbf{X}} {\left[
            e^{
                \cos\left(
                    \mathbf{p}_\mathbf{x}^{r_*}, 
                    \mathbf{p}_{\mathbf{y}'}^{r_*}
                \right) / \tau
            }
            +
            \vmathbb{1}_{\mathbf{y} \neq \mathbf{x}} \cdot
            e^{
                \cos\left(
                    \mathbf{p}_\mathbf{x}^{r_*}, 
                    \mathbf{p}_\mathbf{y}^{r_*}
                \right) / \tau
            }
        \right]}
    },
\end{equation}
and similarly for the transformed data point $\mathbf{x}'$-based positive pair $(\mathbf{x}', \mathbf{x})$, 
the NT-Xent loss $l_{\mathbf{x}'}^{r_*}$ is
\begin{equation}
    \label{eq:simclr}
    l_{\mathbf{x}'}^{r_*} = -\log\frac{
        e^{
            \cos\left(
                \mathbf{p}_{\mathbf{x}'}^{r_*}, 
                \mathbf{p}_\mathbf{x}^{r_*}
            \right) / \tau
        }
    }
    {
        \sum_{\mathbf{y} \in \mathbf{X}} {\left[
            e^{
                \cos\left(
                    \mathbf{p}_{\mathbf{x}'}^{r_*}, 
                    \mathbf{p}_\mathbf{y}^{r_*}
                \right) / \tau
            }
            +
            \vmathbb{1}_{\mathbf{y} \neq \mathbf{x}} \cdot
            e^{
                \cos\left(
                    \mathbf{p}_{\mathbf{x}'}^{r_*}, 
                    \mathbf{p}_{\mathbf{y}'}^{r_*}
                \right) / \tau
            }
        \right]}
    },
\end{equation}
where 
    $\mathbf{p}_\mathbf{x}^{r_*} = P^{r_*}(F^{r_*}(\mathbf{x}))$ and $\mathbf{p}_{\mathbf{x}'}^{r_*} = P^{r_*}(F^{r_*}(\mathbf{x}'))$ 
        are the attribute-agnostic logits computed from the original and transformed data points, respectively;
    $\tau$ 
    controls the contribution proportion between the positive and the negative samples, 
    with a higher $\tau$ value indicating 
    a bigger contribution from the positive samples, 
    and vice versa; and
    $\vmathbb{1}_{\mathbf{y} \neq \mathbf{x}}$ is the indicator function
        $\vmathbb{1}_{\mathbf{y} \neq \mathbf{x}} = 
        \begin{cases}
        1, & \mathbf{y} \ne \mathbf{x}\\
        0, & \text{o.w.}
        \end{cases}$.
The total loss for the preservation branch is thus aggregated across all data points 
\begin{equation}
\label{eq:simclr_all}
L^{r_*} = w^{r_*} \cdot \frac{1}{2N} \sum_{\mathbf{x} \in \mathbf{X}} {( l_\mathbf{x}^{r_*} + l_{\mathbf{x}'}^{r_*})},
\end{equation}
which enables the preservation branch to enforce the underlying representations of 
the transformed data points in $\mathbf{X}'$ to reach maximal agreement with their corresponding origins from $\mathbf{X}$
while maintaining their discriminative characteristics with each other.

\paragraph{Attribute-Specific.}
In addition to the attribute-agnostic case discussed above,
there could also be \textit{attribute-specific} situations,
where the set $R$ of attributes that needs to be preserved
is explicitly defined in advance for the data transformation.
Therefore, we can pretrain the inference models on the original dataset $\mathbf{X}$ like in the suppression branch.
To account for this extra information, $R$,
we formulate the loss function similar to Eq.~\ref{eq:suppression_loss},
except that here we want to maximize the similarities as opposed to minimizing them,
as follows,
\begin{equation}
\label{eq:preservation_loss}
    L^{r} = -w^{r} \cdot \frac{1}{N}\sum_{\mathbf{x} \in \mathbf{X}} {
        \text{sim}^{r}\left(
            \mathbf{x}',
            \mathbf{x}
        \right)
    },
\end{equation}
where $w^{r}$ is the weight for each attribute $r \in R$.

Finally, collecting all the loss terms from the data modifier, 
the suppression branch, 
as well as the attribute-agnostic and attribute-specific components of the preservation branch, 
we have the overall optimization loss
\begin{equation}
\label{eq:total_task}
    L = \sum_{s\in S}{L^{s}} + L_\text{rec} + L^{r_*} + \sum_{r\in R}{L^{r}},
\end{equation}
which is similar to GAN that the data modifier tries to minimax the loss.

%% file: Tex/04_experiments.tex
\section{Experimental Evaluation}
\label{sec:exp}

We next present and discuss our experimental evaluation of \ours
on its ability to perform selective attribute suppression and preservation when carrying out data transformations.

\subsection{Experimental Setup}
We evaluated \ours on three multi-attribute datasets of different domains, namely Adience~\citep{eidinger2014age} for facial images,
AudioMNIST~\citep{becker2018interpreting} for voice recordings,
and PA-HMDB~\citep{wu2020privacy} for video clips.
For all datasets, we converted all their raw data points to feature embeddings via state-of-the-art neural networks as the input $\mathbf{X}$.
Table~\ref{table:model_setup} lists the feature extractors 
used for each dataset, as well as the corresponding feature dimension and the architecture of MLP used in the data modifier.
All our experiments were designed to examine \ours' performance on attribute suppression and preservation quality, 
and not whether it could generate high quality synthetic data.
Therefore, each transformed $\mathbf{x}'$ would stay in the same feature space as its corresponding $\mathbf{x}$.
We next briefly introduce the datasets,
as well as the training and evaluation protocols.
More details can be found in Appx.~\ref{sec:app:data_preprocess}.

\paragraph{Adience.} The Adience image
dataset was originally published to help study 
the recognition of age and gender.
Each image is also associated with a \adienceid.
In total, the dataset used in our experiment contains
1,089 different {\adienceid}s, 
8 age groups, 
and 2 gender classes.
We split the dataset into 2,815 for training and 2,525 for validation.

\paragraph{AudioMNIST.} The AudioMNIST
dataset contains audio recordings of 
spoken digits (0-9) in English from 60 speakers. 
In addition to speakerID and spoken digits, 
the dataset also contain accent and gender attributes.
Therefore, we use AudioMNIST as a 4-attribute dataset,
namely speakerID, spoken digits, accent, and gender,
with 60, 10, 18, and 2 classes, respectively. 
There are 30,000 audio clips in total.
We split the data into 18,000, 6,000, and 6,000 
for training, validation, and testing, respectively.

\paragraph{PA-HMDB.} The PA-HMDB51 dataset is subset of HMDB51 and it contains 6 attributes:
action, skin color, face, gender, nudity, and relationship~\citep{kuehne2011hmdb,wu2020privacy}.
The action label is annotated at video-level while the other non-action attributes are at frame-level.
Nonetheless, PA-HMDB51 only contains about 500 videos, it is only used for evaluation. 
There are 51 different action classes.
The other 5 non-action attributes are all binary.
As there is no training data for the non-action attributes in PA-HDMB, 
we used the VISPR dataset~\citep{orekondy2017towards} 
for training them,
and used HMDB51 for the action attribute. 
\ours was evaluated on both VISPR and PA-HMDB51.

\subimport{./}{Tables/model_arch}
\paragraph{Training and Evaluation Protocols.}
Before we can start training \ours' data modifier $G$, 
we first need to train the models 
for the attributes we intend to suppress
in the suppresion branch,
as well as the attribute-agnostic model 
for the preservation branch,
and potentially also the attribute-specific models 
if the corresponding labels for those attributes are available. 
Subsequently, each of these pre-trained models is used in either the suppression or the preservation branch 
during the training of the data modifier $G$ via optimizing Eq.~\ref{eq:total_task} defined in Sec.~\ref{sec:proposed_methods}. 
Note that during training $G$, the pre-trained models in both branches are fixed without any update.
For evaluation, we use the trained data modifier to generate new data 
and feed them through the pre-trained 
models and examine their performance.
For both Adience and AudioMNIST, 
we use top-1 accuracy as the metric 
and report the results on each attribute. 
For PA-HMDB51, we follow the same practice as PA-HMDB~\citep{wu2020privacy} to aggregate 
the performance of the 5 non-action attributes
via macro-average of classwise mean average precision (cMAP)
and use top-1 accuracy for action. 
More details on the model training can be found in Appx.~\ref{sec:app:model_pretraining}.

\subsection{Comparison to Baselines}

\subimport{./}{Tables/baseline_adience}

First, we examine the effects of each of the loss terms
for each of the datasets. 
Table~\ref{table:adience_results} shows the performance on Adience where \adienceid is targeted for suppresion,
and the others are expected to be preserved.
When only imposing attribute suppression ($L_{\text{rec}}$ and $L^{s_1}$), 
\ours did successfully degrade the performance of \adienceid.
But at the same time, 
the performance on age and gender 
deteriorated significantly. 
When we added the attribute-agnostic loss ($L^{r_*}$), 
\ours still achieved good suppression on \adienceid.
But at the same time it greatly improved the recognition accuracy on age and gender.
Note that, we do not use any information from age and gender attributes during the training.
Moreover, when the age information ($L^{r_1}$) was incorporated in \ours, the performance of age detection on the transoformed dataset almost returned to the same level as that of the  original dataset.
The same trend can be observed for the gender attribute ($L^{r_2}$) as well.

\subimport{./}{Tables/baseline_audiomnist}

\subimport{./}{Tables/baseline_pahmdb}

A similar outcome can also be observed for the AudioMNIST dataset (Table ~\ref{table:audioMNIST_results}), 
where \audiomnistid was treated as the suppression target,
and all the rest of the attributes 
were expected to be preserved.
When \ours only used suppression ($L_{\text{rec}}$ and $L^{s_1}$), 
\audiomnistid score did reduce,
but so did the other attributes.
Adding the attribute-agnostic loss ($L^{r_*}$)
brought significant improvement 
for the performance of all the other three attributes.
Finally, when the attribute information were available ($L^{r_1}, L^{r_2}$ and $L^{r_3}$), 
\ours was able to boost the preservation performance 
to near perfection
without reducing the suppression quality on \audiomnistid.

Table~\ref{table:pahmdb_results} shows the performance of \ours on the video dataset. 
With only suppression ($L_{\text{rec}}$ and $L^{s_1}$),
\ours indeed lowered the cMAP on both VISPR and PA-HMDB.
However, the performance on the action attribute also dropped significantly. 
By adding the attribute-agnostic loss ($L^{r_*}$), 
\ours was able to significantly improve the accuracy of the action attribute 
without utilizing any label information.
We do notice a slight decrease in the performance of other attributes, 
which also aligns with the observation in SPAct~\citep{spact}, 
When the action label ($L^{r_1}$) was incorporated, 
\ours was able to preserve the same action accuracy while suppressing the other attributes.

All experimental results above show that \ours is highly configurable and effective 
even without knowing the downstream tasks in advance.
Moreover, when the information of downstream task is available, \ours can boost the performance on the corresponding specific attributes.

\subsection{Comparison to Other Methods}
As \ours includes both suppression and preservation,
there is limited prior work for direct comparison
since most of them focus on suppression only.
Therefore, we first compared \ours with the heuristic methods,
e.g., perturbation on the original data, 
additive noise in the feature space for both the Adience and AudioMNIST datasets.
For Adience, we further compared \ours to CIAGAN~\citep{CIAGAN_Maximov_2020_CVPR} even though it only performed suppression. 
Lastly, we compared \ours to SPAct
on PA-HMDB. 
More details on the comparisons can be found in Appx.~\ref{sec:app:comparison}.

\subimport{./}{Tables/comp_adience}

Table~\ref{table:adience_comp} shows the comparison on Adience. 
As most methods were designed to suppress \adienceid, 
we configured \ours to suppress \adienceid and preserve the others with attribute-agnostic setting for comparison.
First, adding noise in the feature domain lowered the accuracy on the \adienceid attribute
but it also degraded the performance on age and gender.
On the other hand, for all the methods which manipulate data in the original domain, including Gaussian blurring 
with different kernel sizes and standard deviations,
downsampling 
and upsampling back to the original size,
and obfuscating the various face area where the face is detected by MTCNN~\citep{mtcnn}, 
they led to similar results since they did not learn 
what features should be preserved for downstream tasks.
By modifying faces in the image space,
CIAGAN~\citep{CIAGAN_Maximov_2020_CVPR} was able to 
lower the performance on \adienceid again.
However, it still wasn't capable of preserving any features
for the other attributes.
In contrast,
\ours not only suppressed \adienceid but also preserved much more data utility such that the data could still perform well on age and gender detection tasks,
even though no additional information on age or gender was made available to \ours.

The comparative study results on AudioMNIST are listed in 
Table~\ref{table:audiomnist_comp}.
The heuristic approaches were capable of suppressing \audiomnistid, 
as well as the other attributes indiscriminatively. 
Nonetheless, \ours can retain the other attributes without knowing the downstream tasks 
while still achieving \audiomnistid suppression.
Lastly, comparing to SPAct~\citep{spact} on PA-HMDB\footnote{We compared SPAct based on their Table 8 in the paper, which is the closest to our setting. 
Because the baseline performance is different, 
we show here the relative suppression results.}, 
\ours achieved competitive suppression ratio of cMAP on both datasets (VISPR: 55.6\% vs. 57\% and PA-HMDB: 20.7\% vs. 16\%) for the other 5 non-action attributes.
In addition, \ours also provides the flexibility to configure which attributes to suppress 
while SPAct can only target all five attributes at the same time.

Additional ablation studies of \ours, including suppressing different attributes, effects of loss weights and different similarity measurements, can be found in Appx.~\ref{sec:app:ablation}.

\subimport{./}{Tables/comp_audiomnist}

%% file: Tables/model_arch.tex
\begin{table}[tb!]
    \centering
    \caption{Model configuration for each dataset.}
    \label{table:model_setup}
    \tablevspace
    \begin{adjustbox}{max width=\linewidth}
    \begin{tabular}{llcc}
    \toprule
    \multicolumn{1}{c}{Dataset} & \multicolumn{1}{c}{Feature Extractor} & Feature Dimension & MLP in Data Modifier \\
    \midrule
    Adience &  FaceNet~\citep{mtcnn} & 512 & \textsc{512-256-128-256-512} \\
    AudioMNIST &  HuBERT-L~\citep{hubert} & 1024 & \textsc{1024-256-128-256-1024} \\
    PA-HMDB &  R3D-18~\citep{Tran_2018_CVPR_R2plus1D} & 512 & \textsc{512-256-128-256-512} \\
    \bottomrule
    \end{tabular}
    \end{adjustbox}
\end{table}

%% file: Tables/baseline_adience.tex
\begin{table}[tb!]
    \centering
    \caption{Results on Adience under different configurations. The checkmark ($\checkmark$) denotes that the particular loss is used in the optimization. \adienceid is selected as the suppression target. 
    }
    \tablevspace
        \label{table:adience_results}

    \begin{tabular}{lcccccccc}
    \toprule
    
    \multirow{2}{*}{} & \multicolumn{5}{c}{Loss Configuration} & \multicolumn{3}{c}{Top-1 Accuracy (\%)} \\
    \cmidrule(lr){2-6} \cmidrule(lr){7-9}
        & $L_\text{rec}$ & $L^{s_1}$ & $L^{r_*}$ & $L^{r_1}$ & $L^{r_2}$ &   \adienceid ($s_1$)  & Age ($r_1$) & Gender ($r_2$)  \\
    \midrule
    Original & - & - & - & - & - & 90.8 & 89.1 & 97.4  \\
    \midrule
    \multirow{4}{*}{\ours} 
        & $\checkmark$ & $\checkmark$ & - & - & - & 0.0 & 33.8 & 73.5  \\ 
        & $\checkmark$ & $\checkmark$ & $\checkmark$ & - & - & 0.6 & 78.5 & 95.7   \\
        & $\checkmark$ & $\checkmark$ & $\checkmark$ & $\checkmark$ & - & 0.7 & 86.1 & 95.9   \\
        & $\checkmark$ & $\checkmark$ & $\checkmark$ & $\checkmark$ & $\checkmark$  & 0.6 & 86.9 & 96.7 \\

    \arrayrulecolor{black}\bottomrule
    \end{tabular}

\end{table}

%% file: Tables/baseline_audiomnist.tex
\begin{table}[tb!]
    \centering
    \caption{Results on AudioMNIST under different configurations. 
    The checkmark ($\checkmark$) denotes that the particular loss term is used in the optimization.
    \audiomnistid is selected as the suppression target.
    }
        \label{table:audioMNIST_results}
    \tablevspace
    \begin{adjustbox}{max width=\linewidth}

    \begin{tabular}{lcccccccccc}
    \toprule
    
    \multirow{2}{*}{} & \multicolumn{6}{c}{Loss Configuration} & \multicolumn{4}{c}{Top-1 Accuracy (\%)} \\
    \cmidrule(lr){2-7} \cmidrule(lr){8-11}
        & $L_\text{rec}$ & $L^{s_1}$ & $L^{r_*}$ & $L^{r_1}$ & $L^{r_2}$  & $L^{r_3}$ &   \audiomnistid ($s_1$)  & Digit ($r_1$)  & Accent ($r_2$) & Gender ($r_3$) \\
    \midrule
    Original & - & - & - & - & - & - & 95.6 & 99.8 & 99.3 & 96.5  \\
    \midrule
        \multirow{5}{*}{\ours} & $\checkmark$ & $\checkmark$ & - & - & - & - & 0.0 & 26.2 & 45.3 & 53.5   \\
        & $\checkmark$ & $\checkmark$ & $\checkmark$ & - & - & -  & 1.7 & 67.4 & 68.7 &	88.2   \\
        & $\checkmark$ & $\checkmark$ & $\checkmark$ & $\checkmark$ & - & -  & 1.7 & 99.7 &	68.4 &	80.0   \\
        & $\checkmark$ & $\checkmark$ & $\checkmark$ & $\checkmark$ & $\checkmark$ & - & 1.7 & 99.7 & 95.1 & 86.6  \\
        & $\checkmark$ & $\checkmark$ & $\checkmark$ & $\checkmark$ & $\checkmark$ & $\checkmark$  & 1.7 & 99.6& 95.7 & 98.4   \\

    \arrayrulecolor{black}\bottomrule
    \end{tabular}
    \end{adjustbox}
\end{table}

%% file: Tables/baseline_pahmdb.tex
\begin{table}[tb!]
    \centering
    \caption{Results on VISPR and PA-HMDB under different configurations. The checkmark ($\checkmark$) denotes that the particular loss term is used in the optimization.
    Metrics for the action attribute is Top-1 Accuracy (\%) while cMAP (\%) is used for the other 5 non-action attributes.
    \ours is configured to suppress the non-action attributes.
    }
        \label{table:pahmdb_results}
        \tablevspace
    
    \begin{adjustbox}{max width=\linewidth}

    \begin{tabular}{l cccc c cc}
    \toprule
    
 & \multicolumn{4}{c}{Loss Configuration} & VISPR & \multicolumn{2}{c}{PA-HMDB}  \\
 \cmidrule(lr){2-5} \cmidrule(lr){6-6} \cmidrule(lr){7-8}
        & $L_\text{rec}$ & $L^{s_1}$ & $L^{r_*}$ & $L^{r_1}$ &  Non-action Attrs. ($s_1$) & Action ($r_1$)  & Non-action Attrs. ($s_1$)  \\
    \midrule
    Original & - & - & - & - & 81.8 & 58.7 & 79.7   \\
    \midrule
    \multirow{3}{*}{\ours} & $\checkmark$ & $\checkmark$ & - & - &  41.8 & 12.6 & 70.3  \\
     & $\checkmark$ & $\checkmark$ & $\checkmark$ & - & 36.3 & 52.1 &	63.2 \\
     & $\checkmark$ & $\checkmark$ & $\checkmark$ & $\checkmark$ & 38.6 &	58.0 & 63.4 \\
     \bottomrule

    \end{tabular}
    \end{adjustbox}
\end{table}

%% file: Tables/comp_adience.tex
\begin{table}[tb!]
    \centering
    \caption{Comparison to other methods on Adience. Top-1 accuracy (\%) is reported. Features and Raw Data denote that the modifications are made on the feature level and the image level, respectively.
    }
    \label{table:adience_comp}
    \tablevspace
    \begin{adjustbox}{max width=\linewidth}
    \begin{tabular}{llccc}
    \toprule
    \multicolumn{2}{c}{Method}  & \adienceid ($\downarrow$)  & Age ($\uparrow$) & Gender ($\uparrow$)  \\
    \midrule
    \multirow{2}{*}{Gaussian Noise} & Features, $\sigma=0.5$ & 50.8 & 56.8 & 88.1  \\
                                    & Features, $\sigma=1.0$ & 0.5 & 26.8 & 54.9 \\
    \arrayrulecolor{black!10}\cmidrule(lr){1-5}
    \multirow{2}{*}{Guassian Blur}  & Raw Data, $k=11$, $\sigma=10.0$ & 39.4 & 49.0 & 86.5 \\
                                    & Raw Data, $k=21$, $\sigma=15.0$ & 1.0 & 19.3 & 66.2 \\
    \cmidrule(lr){1-5}
    \cmidrule(lr){1-5}
    \multirow{2}{*}{Downsample} & Raw Data, 8$\times$ & 11.0 & 30.3 & 78.7 \\
                                & Raw Data, 4$\times$ & 79.3 & 74.7 & 94.3 \\
    \cmidrule(lr){1-5}
    \multirow{3}{*}{Obfuscation}    & Raw Data, Face area, ratio $=1.0$ & 0.8 & 17.1 & 59.5 \\
                                    & Raw Data, Face area, ratio $=0.36$ & 4.0 & 29.8 & 81.6 \\
                                    & Raw Data, Face area, ratio $=0.09$ & 57.7 & 64.1 & 93.9\\
    \cmidrule(lr){1-5}
    \multicolumn{2}{l}{CIAGAN}  &   1.1 & 17.8 & 66.9 \\
    \cmidrule(lr){1-5}
    
    \ours & $L_{rec} + L^{s_1} + L^{r_*}$ & 0.6 & 78.5 & 95.7 \\
    
    \arrayrulecolor{black}\bottomrule
    \end{tabular}
    \end{adjustbox}
\end{table}

%% file: Tables/comp_audiomnist.tex
\begin{table}[tb!]
    \centering
    \caption{Comparison with other methods on AudioMNIST. Top-1 accuracy (\%) is reported.}
    \label{table:audiomnist_comp}
    \tablevspace
    \begin{adjustbox}{max width=\linewidth}
    \begin{tabular}{lcccc}
    \toprule
    Method  & \audiomnistid ($\downarrow$)  & Digit ($\uparrow$) & Accent ($\uparrow$)&  Gender ($\uparrow$) \\
    
    \midrule
    White Noise (Raw Data)  & 4.7 & 23.3 & 23.4 & 32.2\\
    Masking (Raw Data)  & 1.8 & 10.6 & 3.9 & 80.0\\
    \ours ($L_{rec} + L^{s_1} + L^{r_*}$) & 1.7 & 67.4 & 68.7 &	88.2 \\
    \bottomrule
    \end{tabular}
    \end{adjustbox}
    
\end{table}

%% file: Tex/05_conclusion.tex
\section{Conclusion}
\label{sec:conclusion}

In this paper, we proposed \ours to selectively suppress attributes while preserving other attributes; 
moreover, with the proposed attribute-agnostic approach, 
the preservation can be achieved without foreseeing the downstream tasks, 
which expands the usability of the proposed algorithm. 
We validated our method in three datasets in different domains, including facial images, voice audio and video clips, and all results showed that our method is promising in attribute suppression and preservation. 
Lastly, we would like to point out that we validated \ours in the feature space rather than 
the original data space;
an interesting future direction,
which is beyond the scope of this paper,
would be
to integrate \ours with generative models, 
such as GAN or diffusion models
to further convert the transformed feature vectors
back into their original representation space (image, audio, etc.).

%% file: Tex/09_appendix.tex
\clearpage
\appendix

\section*{Appendix}
In the appendix, we provided the supplemented materials, including data preprocessing in Sec.~\ref{sec:app:data_preprocess}. 
In Sec.~\ref{sec:app:model_pretraining}, we described model pre-training for contrastive learning, model training for single attribute and \ours. 
We included the ablation studies in Sec.~\ref{sec:app:ablation} and how to generate compared results in Table~\ref{table:adience_comp} and Table~\ref{table:audiomnist_comp} in Sec.~\ref{sec:app:comparison}. 
Our source code and models will be publicly available to help better understand the settings of training and evaluation.

\section{Data preprocessing}
As briefly discussed in the main manuscript, \ours takes a feature vector in and then generates a feature vector instead of operating on the raw data. Therefore, for each dataset, we convert the data into feature vectors via state-of-the-art neural networks and then normalize the vector by its L2-norm.

\label{sec:app:data_preprocess}
\paragraph{Adience.}
We first resize the image into $160\times160$ and normalize the image by the mean and the standard deviation used in the FaceNet~\citep{mtcnn}. Then, we feed the normalized image into FaceNet to get a 512-d feature vector.

\paragraph{AudioMNIST.} 
\label{audiomnistprocess}
The majority of the information in audio signal resides at the beginning, 
and the average length of a waveform is 30,844 samples and 
the upper quartile is 34,380. 
Therefore, we either truncate and 
pad (zeros) the waveform to the length of 30,000 at the end such that the data loader can form them as a batch to speed up the training.
Then, we feed the truncated/padded waveform into Hubert-L~\citep{hubert} 
to get a 1024-d feature vector after performing average pooling on the output of Hubert-L along the time dimension.

\paragraph{VISPR and PA-HMDB.}
R3D-18 is trained with the clip size of $16\times112\times112$ and generates a 512-d feature vector; therefore, we resize the spatial dimension of a video into $112\times112$ and then sample 16 frames (every other frame) out of a video to form a clip.
For VISPR, since it is an image dataset, we generate a 16-frame clip by duplicating the same image and then pass it to R3D-18 to extract features. 
On the other hand, for the action attribute in PA-HMDB, we convert each video into frame-level feature vectors by R3D-18. 
More specifically, for each timestamp, we take its neighboring frame (every other frames) to form a 16-frame clip and then feed it to R3D-18. 
E.g., for a video with 100 frames, we will get 100 512-d feature vectors.
For the other attributes in PA-HMDB, since those labels are image-level instead of video-level, 
we simply assign the labels to the frame-level features extracted above.

\section{Model Training}
\label{sec:app:model_pretraining}

\subsection{Attribute-Agnostic Model Pre-training}
For all datasets, we follow similar practices to train attribute-agnostic models via SimCLR~\citep{pmlr-v119-chen20j-simclr}. 
First, we generate two views of data by different data augmentation in the raw data domain, 
and then the two views of data are passed through the fixed feature extractor to get its feature representation. 
After that, we train a multi-layer perceptron (MLP) as an encoder to learn a generic feature representation over the features extracted by the fixed feature extractor. 
The MLP is composed of two fully-connected layers with the same dimension as the input feature dimension,
and 1-D batch normalization layer and ReLU are added between fully-connected layers.
The trained MLP is served as attribute-agnostic model in \ours. Note that we also adopt the MLP-projector in the contrastive learning to achieve better performance.
We train the model for 100 epochs with temperature 0.07 via the stochastic gradient decent (SGD) optimizer. 
The weight decay is set to 0.0001 and the learning rate starts from 0.05 and then it is annealed with cosine schedule.
In the follow paragraphs, we describe how to generate different views for each dataset.

\paragraph{Adience.} 
To generate two views, we first resize images to $160\times160$ and then randomly flip the image horizontally; 
after that, we randomly perform color jitter via torchvision package with 80\% probability and then convert the image into gray scale with 20\% probability. 

\paragraph{AudioMNIST.}
In this work, we apply two different augmentations \citep{ma2019nlpaug}: random crop on the entire audio with a coverage of 0.4 
and mask with a coverage of 0.5 to each view respectively. 
The crop augmentation removes the selected part from the audio, 
whereas mask substitutes it with zeros. 

We limited the data augmentations used in our methods 
to crop and mask because other augmentation like pitch, loudness, speed, etc. 
would affect the structure of audio signal and potentially won't be able to retain attributes like gender, accent, and age.

\paragraph{PA-HMDB.}
We generate two views of data by following the practice in CVRL~\citep{Qian_2021_CVPR_CVRL}, i.e., for a positive pair, 
two views are extracted from different time instance of a video and the temporal-consistent data augmentation is performed on each view.
The data augmentation is composed of resizing the spatial dimension into $112\times112$, guassian blurring, randomly converting color image into gray image.

\subsection{Attribute-Specific Model Pre-training}
For all attributes in all datasets, we train the attribute-specific model by using the cross-entropy loss against the given label to compute the gradient for all parameters through back-propagation. 
The model contains three fully-connected layers and with the dimension: \textit{input dim-512-256-number of classes}, and the 1-D batch normalization layer and ReLU are added between layers.
We use a batch size of 256 with the AdamW optimizer~\citep{loshchilov2018decoupled_adamw} to train the model for 100 epochs.
The weight-decay is fixed to 0.05 and the initial learning rate is set to 0.01 
and then the learning rate is annealed with the cosine scheduler.

\subsection{\ours Training}
The training on different datasets follows similar settings but with different loss weights.
When training the data modifier $G$ in \ours, all models in the suppression and preservation branches are fixed without any update.
We train all models with 100 epochs with the AdamW optimizer~\citep{loshchilov2018decoupled_adamw} 
The weight-decay is 0.05 and we adopt cosine learning scheduler to anneal the learning rate.
For the loss type, in most of cases, we use cosine similarity measurement 
for the to-be-suppressed attribute and KL divergence for the attribute-specific preservation.
Table~\ref{table:model_training_param} described other training details for different datasets.
\subimport{./}{Tables/appendix_our_training}

\subimport{./}{Tables/appendix_adience}

\section{Ablation Studies}
\label{sec:app:ablation}
In ablation studies, we use the Adience dataset for all experiments 
and we discuss \ours in three perspectives, including suppressing different attributes,
effects of loss weights, 
effects of similarity measurement.

\paragraph{Suppression Target.}
In the main manuscript, we always suppress \adienceid in all experiments;
however, \ours is configurable to suppress any attribute while still preserving others. 
Table~\ref{table:adience_different_targets} shows the results by suppressing different attributes.
Only the performance of the selected attribute is degraded while other attributes are still good.
Note that those results do not include any attribute-specific models in the preservation branch.
The result shows that \ours is flexible to configure to suppress any attribute and preserve others.

\paragraph{Loss Weights.}
Intuitively, the loss weight controls which loss term should be focused on more during the optimization.
In this ablation study, we vary the weights for the attribute-agnostic model 
and the results are shown in Table~\ref{table:adience_simclr_weight}. 
Since $L^{r_*}$ controls how generic the feature representation is, 
the higher weights preserve more generic features;
therefore, the transformed dataset could perform better for all attributes. 
However, when $L^{r_*}$ is 160, the accuracy of \adienceid is also increased 
because the strength of suppression is not strong enough since the weight of $L^{r_*}$ is too high.

\paragraph{Different Similarity Measurement for Suppression.}
We proposed three different measurements for the similarity in the main manuscript.
Those measurements provided similar functionalities conceptually 
but they might work different empirically. 
Table~\ref{table:adience_loss_type} shows the results with different measurements, and all results are close to each other. 
Therefore, for suppression, we use cosine for all experiments.

\section{Compared Results}
\label{sec:app:comparison}
\paragraph{Adience.}
We compared many approaches in Table~\ref{table:adience_comp} and 
here we describe the details for how to generate those results. 
First, for Gaussian noise, we added zero-mean with different standard deviations ($\sigma$) into the original feature vectors to manipulate data. 
For Gaussian blur, downsample and obfuscation are all performed in the raw data domain,
and then the modified data are passed through FaceNet to get the feature representation.
For Gaussian blur, we apply zero-mean with various standard deviations ($\sigma$) with different kernel sizes ($k$) to blur the image.
For downsample, we downsample the data with different ratios and then upsample it back to original size. 
Lastly, for obfuscation, we use MTCNN to detect the location of the face and then remove the face region with different ratios.

On the other hand, for CIAGAN~\citep{CIAGAN_Maximov_2020_CVPR}, 
we first followed CIAGAN's method to pre-extract the masked face and the facial landmark information for the Adience dataset by using the Dlib-ml library~\citep{king2009dlib}. 
And then, the CIAGAN model takes in the Adience images, their landmarks, masked faces and the desired target.

After we obtain the transformed Adience images, we use the same procedure as ours for evaluation: using FaceNet~\citep{mtcnn} 
to extract the feature vector of an image.

\paragraph{AudioMNIST.}
We compared two methods in AudioMNIST, including adding white noise and masking out a portion of waveform based on the nlpaug library~\citep{ma2019nlpaug}.
We use the default parameter for white noise and set the masking ratio to 50\% of the waveform.

%% file: Tables/appendix_our_training.tex
\begin{table}[tb!]
    \centering
    \caption{Model training settings for each dataset in Table~\ref{table:adience_results}, \ref{table:audioMNIST_results} and \ref{table:pahmdb_results}.}
    \label{table:model_training_param}
    \tablevspace
    \begin{adjustbox}{max width=\linewidth}
    \begin{tabular}{ccccl}
    \toprule
    \multicolumn{1}{c}{Dataset} & \multicolumn{1}{c}{\#GPUs} & Batch Size per GPU & Learning Rate & Loss Weights \\
    \midrule
    \multirow{2}{*}{Adience} &  \multirow{2}{*}{1} & \multirow{2}{*}{64} & \multirow{2}{*}{0.00125} & $w_{\text{rec}}=1$, $w^{s_1}=5$, $h^{s_1}=1$ \\
    &&&& $w^{r_*}=40$, $w^{r_1}=1$, $w^{r_2}=1$\\
    \arrayrulecolor{black!10}\cmidrule(lr){1-5}
    \multirow{2}{*}{AudioMNIST} & \multirow{2}{*}{4} & \multirow{2}{*}{128} & \multirow{2}{*}{0.01} & $w_{\text{rec}}=10$, $w^{s_1}=0.001$, $h^{s_1}=0.1$ \\
    &&&& $w^{r_*}=15$, $w^{r_1}=0.01$, $w^{r_2}=0.001$, $w^{r_3}=0.001$\\
    \arrayrulecolor{black!10}\cmidrule(lr){1-5}
    \multirow{2}{*}{PA-HMDB} &  \multirow{2}{*}{4} & \multirow{2}{*}{128} & \multirow{2}{*}{0.01} & $w_{\text{rec}}=1$, $w^{s_1}=1$, $h^{s_1}=0.1$ \\
    &&&& $w^{r_*}=1000$, $w^{r_1}=50$\\
    \arrayrulecolor{black}
    \bottomrule
    \end{tabular}
    \end{adjustbox}
\end{table}

%% file: Tables/appendix_adience.tex
\begin{table}[tb!]
    \centering
    \caption{Results on Adience with different suppressed attribute, 
    the experiments are completed under the loss setting of $L_{rec} + L^{s_1} + L^{r_*}$,
    where $s_1$ is the suppressed attribute.
    }
    \tablevspace
    \label{table:adience_different_targets}
    \begin{tabular}{l c ccc}
    \toprule
    \multirow{2}{*}{} & \multirow{2.5}{*}{Suppressed Attribute} & \multicolumn{3}{c}{Top-1 Accuracy (\%)} \\
    \cmidrule(lr){3-5}
        &  &  \adienceid  & Age  & Gender \\
    \midrule
    \multirow{3}{*}{\ours} 
        & \adienceid & 0.6 & 78.5 & 95.7  \\ 
        & Age & 81.2 & 13.7 & 94.5   \\
        & Gender & 77.2 & 75.3 & 8.3   \\
        
    \arrayrulecolor{black}\bottomrule
    \end{tabular}
\end{table}

\begin{table}[tb!]
    \centering
    \caption{Results on Adience with different weights on $L^{r_*}$, the experiments are completed under the loss setting of $L_{rec} + L^{s_1} + L^{r_*}$. We used $w^{r*}=40$ in our main results.
    }
    \tablevspace
    \label{table:adience_simclr_weight}
    \begin{tabular}{l c ccc}
    \toprule
    \multirow{2}{*}{} & \multirow{2.5}{*}{$w^{r_*}$} & \multicolumn{3}{c}{Top-1 Accuracy (\%)} \\
    \cmidrule(lr){3-5}
        &  &  \adienceid  & Age  & Gender  \\
    \midrule
    \multirow{5}{*}{\ours} 
        & 10 & 0.0 & 69.4 & 73.5  \\ 
        & 20 & 0.0 & 75.0 & 94.8   \\
        & 40 & 0.6 & 78.5 & 95.7   \\
        & 80 & 3.2 & 82.1 & 96.4   \\
        & 160 & 13.5 & 83.5 & 96.4   \\
    \arrayrulecolor{black}\bottomrule
    \end{tabular}
\end{table}

\begin{table}[tb!]
    \centering
    \caption{Results on Adience with different similarity measurements in $L^{s_1}$, the experiments are completed under the loss setting of $L_{rec} + L^{s_1} + L^{r_*}$. $s_1$ is \adienceid.
    }
    \tablevspace
    \label{table:adience_loss_type}
    \begin{tabular}{l c ccc}
    \toprule
    \multirow{2}{*}{} & \multirow{2.5}{*}{Similarity} & \multicolumn{3}{c}{Top-1 Accuracy (\%)} \\
    \cmidrule(lr){3-5}
        &  &  \adienceid  & Age  & Gender  \\
    \midrule
    \multirow{3}{*}{\ours} 
        & Cosine & 0.6 & 78.5 & 95.7   \\
        & KL divergence & 0.2 & 76.2 & 95.8   \\
        & CE & 0.1 & 76.7 &	94.9 \\
    \arrayrulecolor{black}\bottomrule
    \end{tabular}
\end{table}